%% file: main.tex
\documentclass[letterpaper]{article} 

\usepackage{aaai2027}
\nocopyright

\usepackage[hyphens]{url}  
\usepackage{graphicx} 
\usepackage{natbib}  
\usepackage{caption} 
\usepackage{algorithm}
\usepackage{algorithmic}
\usepackage{booktabs}
\usepackage{amsmath,amssymb,amsfonts,bm}

\usepackage{multirow}
\usepackage{makecell}
\usepackage{array}
\usepackage[table]{xcolor}

\newcommand{\method}{CoSAG}

\DeclareMathOperator*{\argmax}{arg\,max}

\title{\method{}: Compact Semantic Anchor Gaussians via Training-Free \\ Rate–Distortion Coding}

\author{
    Yuang Jia\textsuperscript{\rm 1},
    Jinlong Wang\textsuperscript{\rm 1},
    Junhong Lin\textsuperscript{\rm 1},
    Ruiting Dai\textsuperscript{\rm 2},
    Wei Gao\textsuperscript{\rm 1}\corresponding
}
\affiliations{
    \textsuperscript{\rm 1}SECE, Peking University \quad 
    \textsuperscript{\rm 2}University of Electronic Science and Technology of China \\
    yuangjia8@gmail.com\quad gaowei262@pku.edu.cn\\
    \textcolor[HTML]{0D47A1}{\url{https://github.com/YuangJia/CoSAG}}
}

\begin{document}

\maketitle

\begin{abstract}
Open-vocabulary 3D scene understanding is commonly achieved by embedding 2D vision-language
features such as CLIP into a 3D Gaussian Splatting scene, turning it into a text-queryable semantic
field. However, attaching a high-dimensional feature to each of millions of Gaussians inflates a single scene
to gigabytes, which makes storage and deployment the real bottleneck of these fields. Existing
compact methods each learn and ship a per-scene codec, an autoencoder, a quantized codebook, or a
distilled feature field, entangling field construction with field storage and never compressing the
per-Gaussian assignment that holds the bulk of the cost. We argue that construction and storage
should be decoupled, and that storage is a rate--distortion problem over the per-Gaussian binding to
a small anchor table, a structure no prior open-vocabulary method compresses. We present \method{},
which constructs the field without any per-scene training through a closed-form transmittance-weighted
lift, spatially grounded semantic anchors, and multi-view denoising, and stores it with a spatially
predictive entropy coder that ships no decoder. Because the anchors are spatially grounded, the
binding is predictable and therefore highly compressible. The transmittance-weighted lift and
multi-view denoising yield a clean, view-consistent assignment, so the entropy coder spends almost
no rate on correcting noise and instead codes only the residual against its spatial prediction.
\method{} reaches sub-megabyte storage while matching or exceeding the state of the art across the
2D-rendered, 3D-selection, and dense-LSeg protocols, reducing field size by $37$ to $76\times$
relative to LangSplatV2 at higher accuracy.
\end{abstract}

\input{section/introduction}

\input{section/related_work}

\input{section/method}

\input{section/experiment}

\section{Conclusion}
In this paper, we recast the storage of an open-vocabulary semantic Gaussian field as a
rate--distortion problem over the per-Gaussian binding, decoupled from field construction, a
structure that no prior open-vocabulary method compresses. Our \method{} constructs the field with no per-scene optimization, through a closed-form lift,
spatially grounded semantic anchors with joint spatial-semantic binding, multi-view denoising,
and a contrast-selected hierarchy. It then stores the field with a spatially predictive coder
that pairs a deterministic per-level projection of the anchor table with Morton-ordered,
cross-level coding of the binding, shipping no per-scene decoder. The spatial grounding of the anchors is what makes the binding predictable
and therefore compressible, driving the field into the sub-megabyte regime while matching or
exceeding optimization-based methods across the 2D-rendered, 3D-selection, and dense-LSeg
protocols. Its limitation is a reliance on rendered depth to ground the anchors, which
weakens where geometry is unreliable; removing this dependence and coding the binding under
richer spatial contexts are promising directions for future work.

\bibliography{cosag}
\clearpage
\input{section/appendix}
\end{document}

%% file: section/introduction.tex
\section{1~~Introduction}

Open-vocabulary 3D scene understanding, which segments and localizes arbitrary
text-queried objects in a reconstructed scene, underpins intelligent
robotics~\citep{openscene,2023conceptfusion}, autonomous driving~\citep{gaussiandwm}, and augmented
reality~\citep{surveyvrar}. Built on 3D Gaussian Splatting (3DGS)~\citep{kerbl3dgs}, the
prevailing approach distills the features of an image-text model such as
CLIP~\citep{clip} into the Gaussian primitives, so that the reconstructed scene answers
free-form text queries~\citep{langsplat,feature3dgs,opengaussian,drsplat,zuo2025fmgs,li2026langsurf,jiao2025clip,li2025instancegaussian,zhou2026ea3d}.
Such fields deliver strong open-vocabulary capability, but endowing every Gaussian with
semantics is costly: a CLIP feature is high-dimensional and a scene holds millions of
Gaussians, so a single semantic field reaches the gigabyte scale. It dwarfs the geometry it
annotates and becomes the principal obstacle to storing, transmitting, and deploying these
fields.

\begin{figure}[t]
\centering
\includegraphics[width=1\columnwidth]{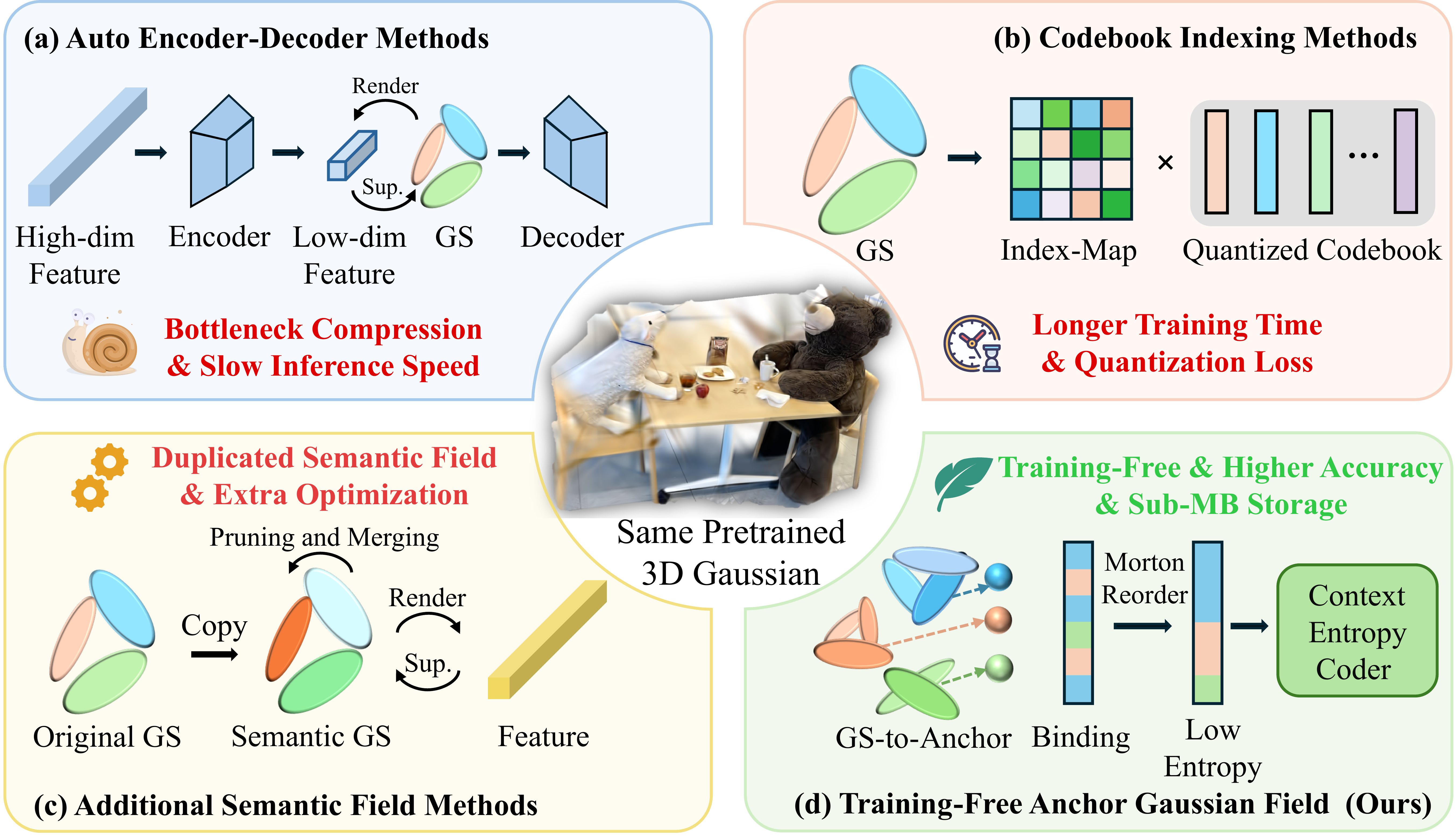}
\caption{\textbf{Compressing an open-vocabulary semantic field on the same pretrained 3D
Gaussians.} (a) Autoencoders suffer bottleneck loss and slow decoding. (b) Codebook indexing
incurs extra training and quantization loss. (c) Additional-field methods train a separate
semantic field, leaving two disjoint Gaussian fields whose appearance and semantics must be
rendered separately. Our \method{} (d) is training-free: it binds Gaussians to spatially
grounded anchors and entropy-codes the resulting binding, reaching sub-megabyte storage with
no decoder and higher accuracy.}
\label{fig:teaser}
\vspace{-15pt}
\end{figure}

Compressing this field has become the central concern, and three lines of work have tackled
it, each shrinking the per-Gaussian feature at a cost (Fig.~\ref{fig:teaser}(a-c)). Early
methods~\citep{langsplat,df3dgs} fit a scene-specific autoencoder and keep only its latent on
each Gaussian, but the bottleneck discards fine-grained semantics and every query must invoke
the decoder. Codebook methods~\citep{legaussians,langsplatv2} learn a shared dictionary and
store a per-Gaussian index into it, yet optimizing that dictionary lengthens training and
quantizing rich semantics into it erodes accuracy. A third line~\citep{df3dgs,cf3} distills a
dedicated semantic field on a separate set of Gaussians, restoring accuracy but leaving two
disjoint fields that must be rendered separately. Without a per-Gaussian correspondence, any
task needing both appearance and semantics, such as 3D selection and scene editing, becomes
cumbersome. These paradigms share two root causes. \textbf{First, each obtains its compression
through optimization, so construction is slow and leaves a scene-specific decoder, codebook, or
feature field stored alongside the scene}. \textbf{Second, each compresses every Gaussian
independently, overlooking that nearby Gaussians on the same object almost always carry
identical semantics, a large and obvious redundancy left untouched}.

We argue that both limitations stem from a common conflation: prior methods fold field construction and field storage into a single trained module. Separating them removes both limitations. For construction, lifting CLIP onto a frozen 3DGS needs no optimization at
all, since the feature minimizing a per-Gaussian, transmittance-weighted reconstruction error
is simply the transmittance-weighted average of the pixel features it renders to, available
in closed form from a single pass. For storage, the lifted field decomposes into a
small anchor table and a per-Gaussian binding to its anchor, so its entire cost lies in
encoding that binding, which is in turn highly compressible: spatially adjacent Gaussians
overwhelmingly share one anchor, turning storage into a classical rate-distortion problem.
Our paradigm follows (Fig.~\ref{fig:teaser}(d)): construct the field without any training,
and store it by compressing the binding rather than the per-Gaussian features, so no decoder
is ever stored or executed.

We realize this paradigm as \method{} (Compact Semantic Anchor Gaussians).
\textbf{Training-Free Anchor Construction} performs the closed-form lift, extracts spatially
grounded anchors, and binds every Gaussian by joint spatial-semantic proximity, removing
the optimization behind the first limitation while yielding the spatially coherent binding
that compression later exploits. Since discarding optimization risks accuracy,
\textbf{Hierarchical Anchor Denoising} refines each anchor with its multi-view support and
selects the semantic granularity per query by relevancy contrast. Finally,
\textbf{Rate--Distortion Binding Compression} addresses the second limitation, compacting the anchor table with a deterministic projection and entropy-coding the binding along a space-filling curve, reaching sub-megabyte storage with no decoder.

We evaluate \method{} on LERF~\citep{lerf}, 3D-OVS~\citep{3dovs}, and
Replica~\citep{straub2019replica} across four tasks: 2D open-vocabulary segmentation, 2D
localization, 3D segmentation, and scene editing. Without any per-scene training, \method{}
compresses the semantic field into the sub-megabyte regime while surpassing the state of the
art under both the 2D-rendered and 3D-selection protocols. Our main contributions are
summarized as follows.

\begin{itemize}
\item We show that existing open-vocabulary semantic fields conflate field construction with
field storage, and recast storage as rate-distortion coding of the per-Gaussian binding, a
structure that no prior method compresses.
\item We propose \method{}, a fully training-free framework that builds the field through a
closed-form lift over spatially grounded anchors and stores it with a spatially predictive
entropy coder, retaining no decoder.
\item By keeping anchors bound to the original Gaussians, \method{} serves the 2D-rendered
protocol, the 3D-selection protocol, and scene editing from one representation, attaining
state-of-the-art accuracy under both protocols at sub-megabyte storage.
\end{itemize}

%% file: section/related_work.tex
\section{2~~Related Work}
\paragraph{Open-Vocabulary 3D Gaussian Semantic Fields.}
Early methods embed CLIP~\cite{clip} features into radiance fields such as LERF~\cite{lerf,zhi2021place},
whose implicit rendering is slow to train and query, so the explicit
3DGS~\cite{kerbl3dgs} has since become the standard backbone, distilling 2D features from
CLIP or LSeg~\cite{lseg} over SAM~\cite{sam} regions into the Gaussians. These methods differ
mainly in how each Gaussian stores its feature: LangSplat~\cite{langsplat} compresses
SAM-region features at three granularities with a scene-specific autoencoder,
Feature-3DGS~\cite{feature3dgs} rasterizes full high-dimensional features at a heavy memory
cost, and LEGaussians~\cite{legaussians} quantize them into a discrete codebook. Others
target 3D understanding rather than 2D rendering: OpenGaussian~\cite{opengaussian} learns
instance features for direct 3D selection, Gaussian Grouping~\cite{gaussiangrouping} assigns
view-consistent identities, and the training-free methods~\cite{drsplat,marrie2025ludvig} register features
onto frozen Gaussians. Accuracy has improved throughout, yet attaching a high-dimensional
feature to each of millions of Gaussians inflates a scene to the gigabyte scale, which makes
compression unavoidable.

\begin{figure*}[ht]
\centering
\includegraphics[width=1\textwidth]{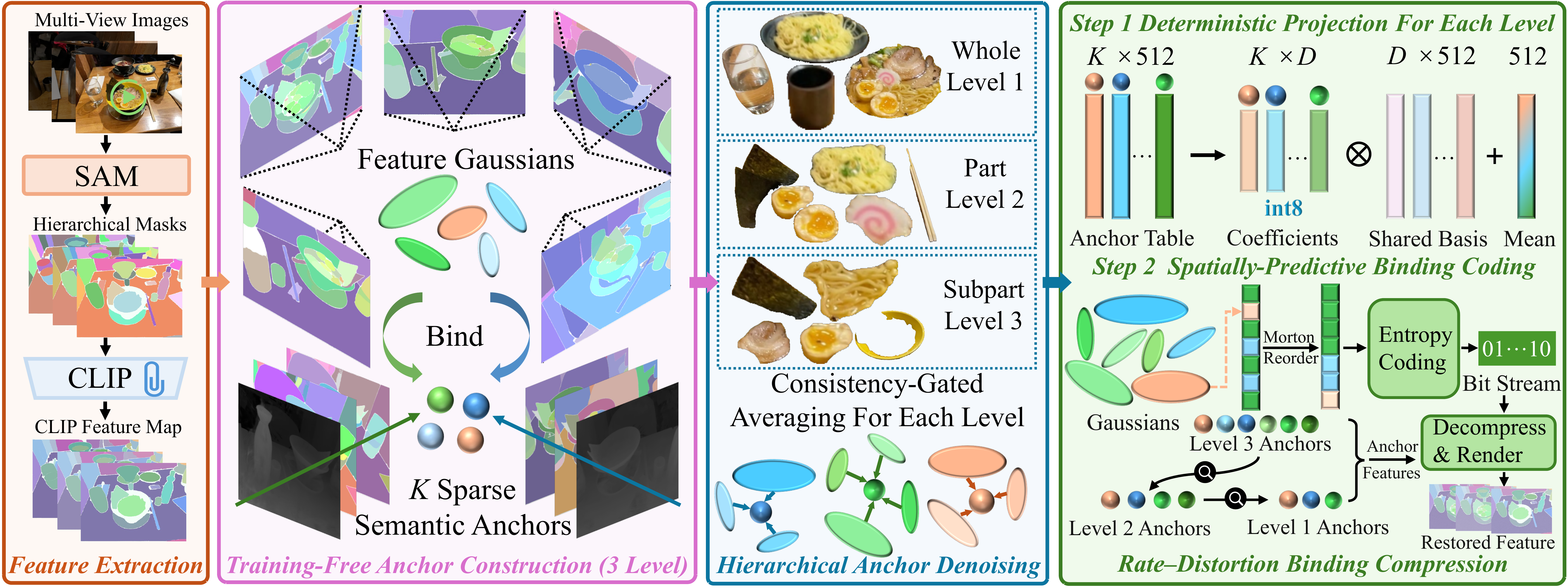}
\vspace{-10pt}
\caption{\textbf{Overview of CoSAG.} We lift per-pixel CLIP features over SAM regions in closed
form onto a frozen 3DGS to obtain per-Gaussian features. Each Gaussian is then bound to a spatially
grounded anchor by joint spatial and semantic proximity, and the anchors are denoised across views
into three granularities. For storage, a shared basis compresses the anchor table while a spatially
predictive coder handles the binding; Morton reordering and parent-pointer sharing drive all three
levels below one megabyte with no shipped decoder.}
\vspace{-10pt}
\label{fig:pipeline}
\end{figure*}

\paragraph{Compression of 3D Semantic Fields.}
The geometry and appearance of 3DGS have been compressed extensively, from anchor-based
Scaffold-GS~\cite{scaffoldgs} to context-model coders such as HAC~\cite{hac} and
ContextGS~\cite{contextgs}, but that line is orthogonal to semantics and has never been
extended to the semantic field. Work that does compress the semantic field still retains a
per-scene learned module: LangSplat~\cite{langsplat} stores a latent and its decoder,
LEGaussians~\cite{legaussians} and LangSplatV2~\cite{langsplatv2} store per-Gaussian indices
over a learned dictionary, and CF3~\cite{cf3} and DF-3DGS~\cite{df3dgs} optimize a separate
semantic field. None of them compress the per-Gaussian assignment itself, where the bulk of
the storage lies. Two recent methods push efficiency further but leave this gap open.
SCOUP~\cite{scoup} learns a 2D sparse code per region and uplifts it with top-K filtering,
yet still trains a per-scene codebook and keeps an uncompressed sparse code on every
Gaussian. The concurrent LightSplat~\cite{lightsplat} injects a two-byte mask index without
training, yet operates at a single object level rather than a multi-granularity hierarchy,
backs its indices with an uncompressed feature table, and does not compress the index field further. \method{} is the first to treat this per-Gaussian assignment as a signal and code it directly. Training-free, multi-level, and entropy-coded along a space-filling curve, it reaches sub-megabyte storage while serving both the 2D-rendered and 3D-selection protocols.

%% file: section/method.tex
\section{3~~Method}

\subsection{3.1~~Preliminaries}
\vspace{-1pt}
\paragraph{Frozen geometry and rendering weights.} The input is a set of posed images
$\{(I_v,\pi_v)\}_{v=1}^{V}$ and a pretrained 3DGS model $\mathcal{G}=\{g_i\}_{i=1}^{N}$ that
we keep frozen throughout, each Gaussian $g_i$ carrying a center $\bm\mu_i\in\mathbb{R}^3$, a
covariance $\bm\Sigma_i$, and an opacity $o_i$. Under camera $\pi_v$, alpha compositing
assigns Gaussian $i$ the blending weight at pixel $p$:
\begin{equation}
w_{i,p}^{v}=\alpha_{i,p}^{v}\!\!\prod_{j\prec_p i}\!\bigl(1-\alpha_{j,p}^{v}\bigr),
\quad \alpha_{i,p}^{v}=o_i\,\mathcal{N}_i(p;\pi_v),
\end{equation}
where $\prec_p$ is the front-to-back order and $\mathcal{N}_i$ is the $2$D Gaussian projected
from $(\bm\mu_i,\bm\Sigma_i)$ under $\pi_v$. Our goal is to attach a semantic embedding
$\mathbf f_i$ to every Gaussian, so that the same compositing renders the per-pixel feature
$\hat{\bm\phi}_p=\sum_i w_{i,p}^{v}\mathbf f_i$. These weights are all we borrow from
rendering; since $\mathcal{G}$ is fixed, we precompute them once.
\vspace{-2pt}
\paragraph{Feature extraction.}
Following the LangSplat series~\cite{langsplat,langsplatv2}, we prepare per-view supervision
from off-the-shelf $2$D models. For each view we run SAM~\cite{sam} at $L{=}3$ granularities,
from whole objects at level $1$ to subparts at level $3$, encode every masked region with
CLIP~\cite{clip}, and propagate its $\ell_2$-normalized embedding to the region's pixels,
resolving overlaps in favor of the smaller. This yields, per level, a dense feature map
$\Phi^{v,\ell}=\mathrm{CLIP}(\mathrm{SAM}^{\ell}(I_v))\in\mathbb{R}^{512\times H\times W}$,
the target of our closed-form lift.
\vspace{-2pt}
\paragraph{Open-vocabulary relevancy.}
Given a rendered per-level feature $\hat{\bm\phi}_p^{\ell}$ and a query embedded as
$\bm\tau_q$, we follow~\cite{lerf,langsplat} and compute the standard relevancy at temperature
$T$ against their canonical negatives $\{\bm\nu_j\}$:
\begin{equation}
R_q^{\ell}(p)=\min_{j}
\frac{\exp(\langle\hat{\bm\phi}_p^{\ell},\bm\tau_q\rangle/T)}
{\exp(\langle\hat{\bm\phi}_p^{\ell},\bm\tau_q\rangle/T)
+\exp(\langle\hat{\bm\phi}_p^{\ell},\bm\nu_j\rangle/T)},
\end{equation}
so a pixel scores high only when the query dominates every negative. Sec.~3.4 evaluates it
per level.

\subsection{3.2~~Notation and Overview}
We separate the two questions that prior methods entangle: \textbf{(Q1) how to construct a per-Gaussian semantic field whose rendered relevancy matches the query}, and \textbf{(Q2) how to store it in the fewest bits}. We reduce Q1 to constructing a compact anchor table $\mathbf{B}^{\ell}\in\mathbb{R}^{K_\ell\times 512}$ together with a per-Gaussian binding $\mathbf{a}^{\ell}\in\{1,\dots,K_\ell\}^{N}$, with $K_\ell\!\ll\!N$, so that the reconstructed
field is the gather $\tilde{\mathbf f}^{\ell}_i=\mathbf b^{\ell}_{a^{\ell}_i}$. Q2 then becomes rate--distortion coding of $\mathbf{B}^{\ell}$ and $\mathbf{a}^{\ell}$, rather than of all $N$ per-Gaussian features. Fig.~\ref{fig:pipeline} shows the pipeline: taking the feature maps
$\{\Phi^{v,\ell}\}$ as input, \method{} runs three training-free stages that answer Q1 in closed form, restore the accuracy that abandoning optimization would cost, and answer Q2 without any learned module.

\subsection{3.3~~Training-Free Anchor Construction}
\paragraph{Closed-form transmittance-weighted lift.}
Optimization-based fields recover $\{\mathbf f_i\}$ from the rendering objective
$\sum_{v,p}\lVert\sum_i w_{i,p}^{v}\mathbf f_i-\Phi^{v,\ell}_p\rVert_2^2$ with thousands of
differentiable-rendering steps, since its normal equations couple all $N$ Gaussians through
the Gram matrix of the compositing weights. We instead minimize the decoupled surrogate
$\sum_{v,p}\sum_i w_{i,p}^{v}\lVert\mathbf f_i-\Phi^{v,\ell}_p\rVert_2^2$, which removes that
coupling and admits, for each Gaussian independently, a unique closed-form minimizer, the
transmittance-weighted average of the pixels it renders to:
\begin{equation}
\label{eq:lift}
\mathbf f^{\ell}_i=\frac{\sum_{v}\sum_{p} w_{i,p}^{v}\,\Phi^{v,\ell}_{p}}
{\sum_{v}\sum_{p} w_{i,p}^{v}+\varepsilon},
\qquad
\hat{\mathbf f}^{\ell}_i=\frac{\mathbf f^{\ell}_i}{\lVert\mathbf f^{\ell}_i\rVert+\varepsilon},
\end{equation}
where $\varepsilon>0$ guards Gaussians with empty support. Both sums accumulate in a single
gradient-free pass over the views, removing the per-scene training that Q1 otherwise demands.

\paragraph{Spatially grounded semantic anchors.}
Storing a $512$-dimensional vector per Gaussian dominates the budget, so we bind the $N$
Gaussians to $K_\ell\!\ll\!N$ anchors forming a learning-free scene vocabulary, each carrying a
seed feature $\mathbf b^{\ell,\mathrm{seed}}_k$. Every SAM region, averaged over the views in
which it appears, seeds one anchor with its $\ell_2$-normalized embedding, grounded in 3D by
back-projecting its pixels through the depth rendered from $\mathcal{G}$; the same
back-projection on a strided grid emits sampling points, each labeled by its anchor. Regions
with no CLIP response yield zero-feature rows that act as background anchors.
Eq.~\eqref{eq:lift} simultaneously yields, at no extra cost, the transmittance-weighted
variance $\mathbf s^{\ell}_i$ of those same features. The $\rho\!=\!10^{-4}$ fraction of Gaussians with the largest $\lVert\mathbf s^{\ell}_i\rVert$
is too view-inconsistent for any region embedding to represent, so each becomes a singleton
anchor at its own $\bm\mu_i$, seeded with its own $\hat{\mathbf f}^{\ell}_i$, emitting no
sampling points and serving no other Gaussian. They add only $\rho N$ rows, yet the rarest primitives survive
intact.

\paragraph{Joint spatial--semantic binding.}
The remaining Gaussians are bound in two stages, a spatial restriction followed by a semantic
gate. For Gaussian $i$ at $\bm\mu_i$, a 3D nearest-neighbor query over the sampling points
returns the $\kappa$ closest, whose labels form the candidate set $\mathcal{A}_\kappa(i)$. A
cosine threshold $\theta$ then retains $\mathcal{A}^{\theta}_\kappa(i)=\{k\in
\mathcal{A}_\kappa(i):\langle\hat{\mathbf f}^{\ell}_i,\mathbf b^{\ell,\mathrm{seed}}_k\rangle
\ge\theta\}$, among which Gaussian $i$ takes the most similar anchor:
\begin{equation}
a^{\ell}_i=\argmax_{k\in\mathcal{A}^{\theta}_{\kappa}(i)}
\langle\hat{\mathbf f}^{\ell}_i,\mathbf b^{\ell,\mathrm{seed}}_k\rangle,
\end{equation}
falling back, when $\mathcal{A}^{\theta}_\kappa(i)=\emptyset$, to the anchor of its nearest
background sampling point, where no gate applies since the feature is null. Coupling locality
with semantic agreement avoids the cross-object leakage of a spatial-only nearest neighbor
and the geometric incoherence of a semantic-only one. Since neighboring Gaussians share an
anchor, $\mathbf a^{\ell}$ is piecewise constant over 3D regions, which Sec.~3.5 exploits.

\subsection{3.4~~Hierarchical Anchor Denoising}
Dropping optimization risks accuracy: a region embedding from a single view is noisy under
occlusion, view-dependent appearance, and SAM over-segmentation. We suppress this noise
without enlarging the anchor table.

\paragraph{Multi-view anchor features.}
Let $\mathcal{C}^{\ell}_k=\{i:a^{\ell}_i=k\}$ collect the Gaussians bound to anchor $k$. We
replace its seed with the scatter-mean of their lifted features, which are already unit
vectors, renormalized to the unit sphere:
\begin{equation}
\mathbf b^{\ell}_k=\operatorname{normalize}
\Bigl(|\mathcal{C}^{\ell}_k|^{-1}\!\!\sum_{i\in\mathcal{C}^{\ell}_k}\hat{\mathbf f}^{\ell}_i\Bigr),
\end{equation}
reverting to the seed when the support is empty. Averaging over many Gaussians observed from
many views cancels the single-view noise of any one region embedding, and unlike per-scene
optimization it introduces no learned parameter and leaves the anchor table unchanged.

\paragraph{Hierarchy and contrast-based level selection.}
These steps run independently at each of the $L$ levels, yielding $L$ clean tables, bindings,
and fields. Averaging the levels into one field hurts accuracy, as their relevancy scales are
uncalibrated and noisy fine levels inject spurious peaks. We instead defer the choice to query
time. Each level is rasterized to a relevancy map $R^{\ell}_q$ and blended with its own
$29\!\times\!29$ box average, $\bar R^{\ell}_q=\tfrac12(\mathrm{box}\!*\!R^{\ell}_q
+R^{\ell}_q)$, which attenuates isolated peaks while retaining local detail; we then select the
level with the sharpest peak-to-background contrast:
\begin{equation}
\ell^{\star}_q=\argmax_{\ell}\Bigl(\max_p \bar R^{\ell}_q(p)
-\operatorname{median}_p \bar R^{\ell}_q(p)\Bigr).
\end{equation}
Taking the median rather than the mean keeps the background estimate robust to the queried
object's own pixels, so a fine level carrying one spurious hot pixel is rejected. The
hierarchy thus adapts its granularity to each query.

\subsection{3.5~~Rate--Distortion Binding Compression}
After construction and denoising, the field is $L$ tables $\{\mathbf B^{\ell}\}$ and $L$
bindings $\{\mathbf a^{\ell}\}$, which we compress with no learned module by exploiting the
redundancy specific to each.

\paragraph{Shared-basis feature coding.}
The $K_\ell$ anchor features of a level are highly correlated and span a low-dimensional
subspace, so we shrink each table by a deterministic principal projection. For
$\mathbf{B}\in\mathbb{R}^{K_\ell\times 512}$ with mean $\bm m$ and centered
$\mathbf X=\mathbf B-\mathbf 1\bm m^{\!\top}$, we take the symmetric eigendecomposition of
$\mathbf X^{\!\top}\mathbf X$, keep the top-$D$ eigenvectors $\mathbf U_{D}$, and store:
\begin{equation}
\mathbf Z=\mathbf X\,\mathbf U_{D},\qquad
\hat{\mathbf B}=\operatorname{normalize}
\bigl(\mathbf Z\,\mathbf U_{D}^{\!\top}+\mathbf 1\bm m^{\!\top}\bigr).
\end{equation}
Each anchor becomes a row of $D$ coefficients over a basis and mean shared across the level, so
only $\mathbf Z\in\mathbb{R}^{K_\ell\times D}$ scales with $K_\ell$; we quantize it to int8
under a symmetric per-column scale, while $\mathbf U_{D}$ and $\bm m$ are stored once. We set
$D$ by semantic richness rather than uniformly, giving the coarse whole-object level $D{=}128$
and the finer levels $D{=}32$ and $D{=}16$. Since the decoding map is affine, it commutes with
alpha compositing, so the coefficient field may equivalently be rasterized in dimension $D$.

\paragraph{Spatially predictive binding coding.}
Storage is now dominated by the binding, one of $K_\ell$ symbols per Gaussian. General byte
compressors gain little here, as the binding is a semantic signal rather than a smooth one, yet it
is spatially structured because neighboring Gaussians share an anchor. We therefore reindex the
Gaussians along a 3D Morton (Z-order) curve, after which most consecutive symbols agree and the
stream forms long low-entropy runs. Coding it losslessly with LZMA then approaches the
order-one conditional entropy $H(a^{\ell}_i\mid a^{\ell}_{i-1})$ that the ordering induces,
which already suffices for a single-level field.

\input{tables/table1}

\input{tables/table2}

\paragraph{Hierarchical binding via parent pointers.}
Coding the $L$ bindings independently would multiply this cost, yet a Gaussian's coarse anchor
is nearly determined by its fine one, since the SAM granularities are nested and a fine region
lies inside one coarse region. The chain rule makes the redundancy explicit,
$H(\mathbf a^{1},\dots,\mathbf a^{L})=H(\mathbf a^{L})+\sum_{\ell<L}
H(\mathbf a^{\ell}\mid\mathbf a^{>\ell})$, where nesting drives every conditional term to
nearly zero. We thus code only the finest binding $\mathbf a^{L}$ and replace each residual
term by an anchor-level parent table
$\mathrm{par}_{\ell}(k)=\operatorname{mode}\{\,a^{\ell-1}_i:a^{\ell}_i=k\,\}$, exact but for the few boundary Gaussians whose parent differs.\footnote{Ties in the mode are broken by the
smallest anchor index.} A parent table holds only $K_\ell$ int16 entries, and a coarser binding
follows by the chained lookup $a^{\ell-1}_i=\mathrm{par}_{\ell}(a^{\ell}_i)$, so the full
hierarchy costs little more than one binding. \method{} therefore stores the coded finest
binding, the parent tables, and the per-level $(\mathbf Z,\mathbf U_{D},\bm m)$; decoding is an
entropy decode, a chain of lookups, and one affine map before a gather, with no per-scene
network stored or executed. The same field renders to a 2D relevancy map or is queried directly
in 3D against $\bm\tau_q$, so one representation serves both protocols.

%% file: tables/table1.tex
\begin{table*}[t]
\centering
\caption{\textbf{Quantitative comparison for Open-vocabulary 2D segmentation and language-field storage on LERF-OVS}.
We report mAcc (\%), mIoU (\%) and storage (MB). \textcolor{red}{Red} and \textcolor{orange}{orange} highlight the best and second-best results in each column.}
\vspace{-5pt}
\label{tab:lerf}
\resizebox{\textwidth}{!}{%
\begin{tabular}{l|ccc|ccc|ccc|ccc|ccc}
\toprule
\multirow{2}{*}{\textbf{Method}}
& \multicolumn{3}{c|}{\textbf{Teatime}} & \multicolumn{3}{c|}{\textbf{Ramen}}
& \multicolumn{3}{c|}{\textbf{Figurines}} & \multicolumn{3}{c|}{\textbf{Kitchen}}
& \multicolumn{3}{c}{\textbf{Overall}} \\
& mAcc & mIoU & Stor. & mAcc & mIoU & Stor. & mAcc & mIoU & Stor. & mAcc & mIoU & Stor.
& mAcc & mIoU & Stor. \\
\midrule
LangSplat~\cite{langsplat} (CVPR'24)      & 88.1 & 65.1 & 78.3 & \cellcolor{orange!25}73.2 & 51.2 & 27.5 & 80.4 & 44.7 & 29.6 & \cellcolor{red!25}95.5 & 44.5 & 76.1 & 84.3 & 51.4 & 52.9 \\
Feature-3DGS~\cite{feature3dgs} (CVPR'24) & 77.2 & 58.8 & 1100.0 & 69.8 & 43.7 & 377.7 & 73.4 & 40.5 & 407.4 & 87.6 & 39.6 &  1068.4 & 77.0 & 45.7 & 738.4 \\
LEGaussians~\cite{legaussians} (CVPR'24)  & 79.7 & 32.3 &  71.7 & 69.0 & 20.2 & 26.6 & 57.1 & 23.4 & 28.4 & 63.6 & 22.3 & 69.8 & 67.4 & 24.6 & 49.1 \\
Gaussian Grouping~\cite{gaussiangrouping} (ECCV'24) & 69.5 & 54.0 & 69.0 & 32.4 & 26.4 & 23.9 & 44.6 & 34.6 & 25.7 & 50.0 & 31.3 & 67.1 & 49.1 & 36.6 & 46.4 \\
CF3~\cite{cf3} (ICCV'25)                 & 88.3 & 68.6 & \cellcolor{orange!25}6.1 & 68.5 & 42.3 & \cellcolor{orange!25}2.2 & 74.0 & 48.8 & \cellcolor{orange!25}2.6 & 76.5 & 49.9 & \cellcolor{orange!25}5.9 & 76.8 & 52.4 & \cellcolor{orange!25}4.2 \\
LangSplatV2~\cite{langsplatv2} (NeurIPS'25) & \cellcolor{red!25}93.2 & \cellcolor{orange!25}72.2 & 77.4 & \cellcolor{red!25}74.7 & 51.8 & 26.6 & \cellcolor{orange!25}82.1 & 56.6 & 28.7 & 86.4 & 59.1 & 75.2 & 84.1 & 59.9 & 52.0 \\
GAGS~\cite{peng2026gags} (AAAI'26)               & 88.1 & 60.3 & 69.7 & 69.0 & 46.8 & 24.6 & 78.6 & 53.6 & 26.4 & \cellcolor{orange!25}90.9 & 55.8 & 67.8 & 81.7 & 54.1 & 47.1 \\
SCOUP~\cite{scoup} (arXiv'26)             & \cellcolor{orange!25}91.5 & \cellcolor{red!25}75.3 & 77.5 & \cellcolor{red!25}74.7 & \cellcolor{orange!25}57.6 & 26.7 & \cellcolor{red!25}85.7 & \cellcolor{orange!25}60.4 & 28.8 & \cellcolor{red!25}95.5 & \cellcolor{orange!25}64.4 & 75.3 & \cellcolor{red!25}86.9 & \cellcolor{orange!25}64.4 & 52.1 \\
\midrule
\textbf{CoSAG (Ours)}                     & \cellcolor{red!25}93.2 & \cellcolor{red!25}75.3 & \cellcolor{red!25}1.7 & \cellcolor{red!25}74.7 & \cellcolor{red!25}60.4 & \cellcolor{red!25}0.9 & 
\cellcolor{red!25}85.7 & \cellcolor{red!25}64.7 & \cellcolor{red!25}1.0 & 
\cellcolor{orange!25}90.9 & \cellcolor{red!25}67.3 & \cellcolor{red!25}1.8 & \cellcolor{orange!25}86.1
 & \cellcolor{red!25}66.9 & \cellcolor{red!25}1.4 \\
\bottomrule
\end{tabular}}
\vspace{-5pt}
\end{table*}

%% file: tables/table2.tex
\begin{table}[t]
\centering
\caption{\textbf{Quantitative comparison for overall 2D open-vocabulary segmentation on 3D-OVS}. We report overall mIoU (\%) and storage (MB). \textcolor{red}{Red} and \textcolor{orange}{orange} highlight the best and second-best results in each column.}
\label{tab:3dovs}
\vspace{-3pt}
\setlength{\tabcolsep}{7.5pt}
\renewcommand{\arraystretch}{1.05}
\begin{tabular}{l|cc}
\toprule
\textbf{Method} & mIoU & Stor. \\
\midrule
LangSplat~\cite{langsplat}                 & \cellcolor{orange!25}93.4 & 54.1 \\
Feature-3DGS~\cite{feature3dgs}            & 87.8 & 755.3 \\
LEGaussians~\cite{legaussians}            & 88.5 & 50.2 \\
Gaussian Grouping~\cite{gaussiangrouping} & 87.7 & 48.2 \\
CF3~\cite{cf3}                            & 84.5 & \cellcolor{orange!25}1.7 \\
LangSplatV2~\cite{langsplatv2}            & \cellcolor{red!25}94.6 & 53.2 \\
SCOUP~\cite{scoup}                        & 92.6 & 53.3 \\
\midrule
\textbf{CoSAG (Ours)}                     & \cellcolor{red!25}94.6 & \cellcolor{red!25}0.7 \\
\bottomrule
\end{tabular}
\vspace{-4pt}
\end{table}

%% file: section/experiment.tex
\section{4~~Experiments}

\subsection{4.1~~Datasets and Implementation Details}
\paragraph{Datasets and protocols.}
We evaluate under three protocols. The \textbf{2D-rendered} protocol uses
LERF-OVS~\cite{lerf,langsplat} and 3D-OVS~\cite{3dovs} with their standard segmentation and
localization annotations. The \textbf{3D-selection} protocol adopts the LERF-OVS 3D-object
annotations of OpenGaussian~\cite{opengaussian}. A \textbf{2D-dense} protocol on
Replica~\cite{straub2019replica} is reported in Sec.~4.4. We report localization accuracy, mIoU,
mAcc, mAcc@0.25 for 3D selection, and language-field storage in MB.

\paragraph{Implementation details.}
Following LangSplat~\cite{langsplat}, we extract OpenCLIP ViT-B/16 features over $L{=}3$
SAM~\cite{sam} granularities on a frozen $30$k-iteration 3DGS~\cite{kerbl3dgs}. CoSAG uses no
per-scene optimization. Across all scenes we fix the cosine gate $\tau{=}0.7$, the neighborhood
$\kappa{=}2000$, the promotion fraction $\rho{=}10^{-4}$, the consistency exponent $\gamma{=}2$, a
$29{\times}29$ relevancy box, and per-level dimensions $D{=}128/32/16$. Experiments run on one
48\,GB RTX~4090.

\subsection{4.2~~2D Open-Vocabulary Segmentation and Localization}
Table~\ref{tab:lerf} evaluates CoSAG on LERF-OVS against eight methods that span the autoencoder,
codebook, and distilled-feature paradigms, including the recent state of the art
SCOUP~\cite{scoup}. Without any per-scene training, CoSAG attains the best overall mIoU of $66.9$
and surpasses SCOUP by $2.5$, and its mAcc of $86.1$ is second only to SCOUP. Its three levels
together occupy $1.4$\,MB, over $37\times$ smaller than LangSplatV2~\cite{langsplatv2} and below
the compact CF3~\cite{cf3}. Table~\ref{tab:3dovs} shows the same pattern on 3D-OVS, where CoSAG
ties LangSplatV2 for the best mIoU of $94.6$ while storing its field in $0.7$\,MB, about $76\times$
smaller than LangSplatV2. The margin comes from feature sharing: each anchor represents hundreds of Gaussians, so storage scales with the few thousand anchors rather than the millions of primitives.

\input{tables/table3}

\subsection{4.3~~3D Open-Vocabulary Object Selection}
CoSAG also selects Gaussians directly in 3D by matching a query against the whole-level anchors
and rendering the selected Gaussians. This uses only the coarsest level, so the query compares
against a few hundred to a few thousand anchors and the stored field records a single level. On
LERF-OVS (Table~\ref{tab:3dsel}), CoSAG reaches the best overall mAcc@0.25 of $72.25$ and mIoU of
$47.85$ at $1.35$\,MB, exceeding the concurrent LightSplat~\cite{lightsplat} by $3.93$ mAcc@0.25 with
$2.3\times$ less storage. Anchor aggregation and multi-view denoising make each coarse anchor a
clean prototype, so matching a query against these few anchors already reaches state-of-the-art
selection. The same field also renders in 2D, so one representation leads under both protocols,
whereas autoencoder methods store a latent that is usable only after decoding and cannot be
matched against a query in 3D.

\input{tables/table4}

\subsection{4.4~~Dense Semantic Segmentation on Replica}
To probe generality beyond SAM-region CLIP, we follow the LSeg-based novel-view protocol of
DF-3DGS~\cite{df3dgs} on Replica~\cite{straub2019replica} room\_0. Only the front end changes. A
dense per-pixel LSeg~\cite{lseg} teacher replaces the SAM regions, so anchors come from cosine
$k$-means on the lifted features and each Gaussian binds to its nearest centroid. The lift and the
feature coding stay the same, while the single level needs neither the consistency hybrid nor the
Morton and parent coding. As Table~\ref{tab:replica} shows, CoSAG improves over DF-3DGS by $1.1$
mIoU and $0.3$ pixel accuracy while shrinking the field from $8.31$ to $0.34$\,MB, and it ships no
autoencoder where DF-3DGS spends $4.05$\,MB on one. The construction and storage therefore
transfer intact from CLIP to dense LSeg.

\input{tables/table5}

\vspace{-5pt}
\subsection{4.5~~Efficiency Analysis}
Table~\ref{tab:eff} compares efficiency on LERF-OVS against the open-source methods whose
released code allows construction time, speed, and memory to be measured under an identical
setup. \method{} builds the field in $0.7$ hours without any optimization, the fastest of all
methods and over $4\times$ faster than LangSplatV2. Rendering in the low dimension $D$ rather
than $512$ allows it to reach $30.2$ FPS at $6.0$\,GB, the highest speed and lowest memory while
keeping the smallest field.

\subsection{4.6~~Ablation Studies}
\paragraph{Construction and denoising.}
Table~\ref{tab:ablation} removes one design at a time. A spatial-only binding ignores appearance
and cuts mIoU by $19.8$, while a semantic-only binding ignores 3D locality and drops it by $4.3$.
Removing the multi-view averaging leaves noisy anchors and costs a further $1.4$ mIoU.

\paragraph{Binding coder.}
Table~\ref{tab:coder} adds each coding step. Raw int16 bindings cost $48.0$ bits per Gaussian, and
LZMA on the native order reaches only $31.4$. Morton reordering exposes long runs and cuts this to
$11.0$ bits, and parent-pointer sharing reaches $4.05$ bits, a $12\times$ reduction to $0.72$\,MB.

\begin{figure*}[t]
\centering
\includegraphics[width=1\textwidth]{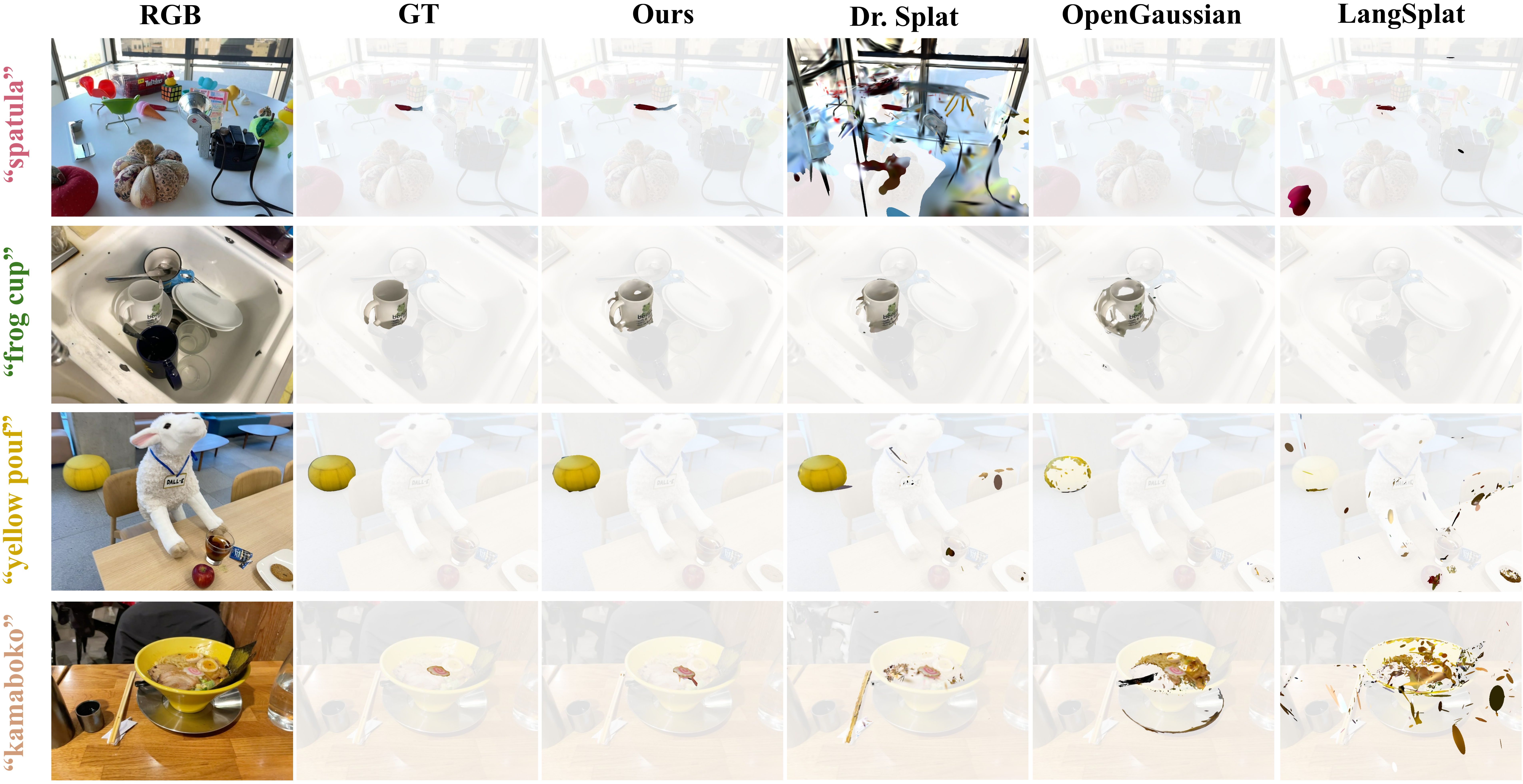}
\vspace{-12pt}
\caption{\textbf{Qualitative 3D object selection on LERF-OVS.} CoSAG selects the Gaussians of the
queried object with coherent boundaries, while competing methods leak into surrounding geometry or
miss parts of the object.}
\label{fig:3dseclect}
\vspace{-12pt}
\end{figure*}

\begin{figure}[t]
\centering
\includegraphics[width=0.9\columnwidth]{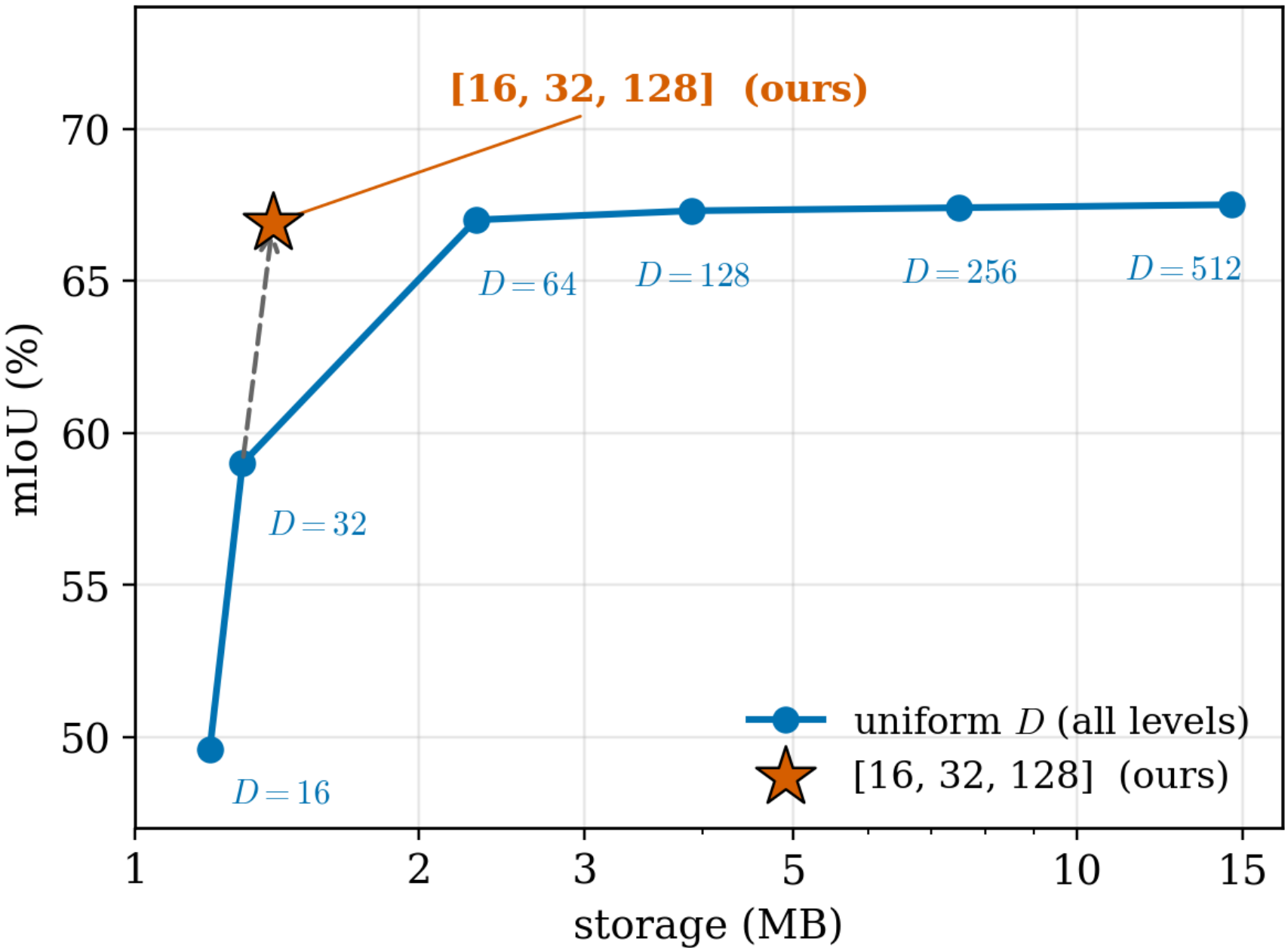}
\vspace{-6pt}
\caption{\textbf{Storage--accuracy trade-off of the projection dimension.} A uniform $D$ (blue)
saturates near $64$ and collapses below it, while our per-level allocation $[16,32,128]$ (orange)
is Pareto-optimal and gains about $+6$ mIoU at equal storage.}
\label{fig:rd}
\vspace{-14pt}
\end{figure}

\paragraph{Per-level projection dimension.}
Figure~\ref{fig:rd} plots mIoU against storage as $D$ varies. A uniform $D$ saturates near $64$ and collapses below it, since too few dimensions underfit the semantically rich coarse level. Our per-level allocation $[16,32,128]$ lies at the upper-left of that curve and gains about $+6$ mIoU at equal storage by spending capacity where it matters, which makes it Pareto-optimal.

\input{tables/table6}
\input{tables/table7}

\subsection{4.7~~Qualitative Results}
\begin{figure}[t]
\centering
\includegraphics[width=1\columnwidth]{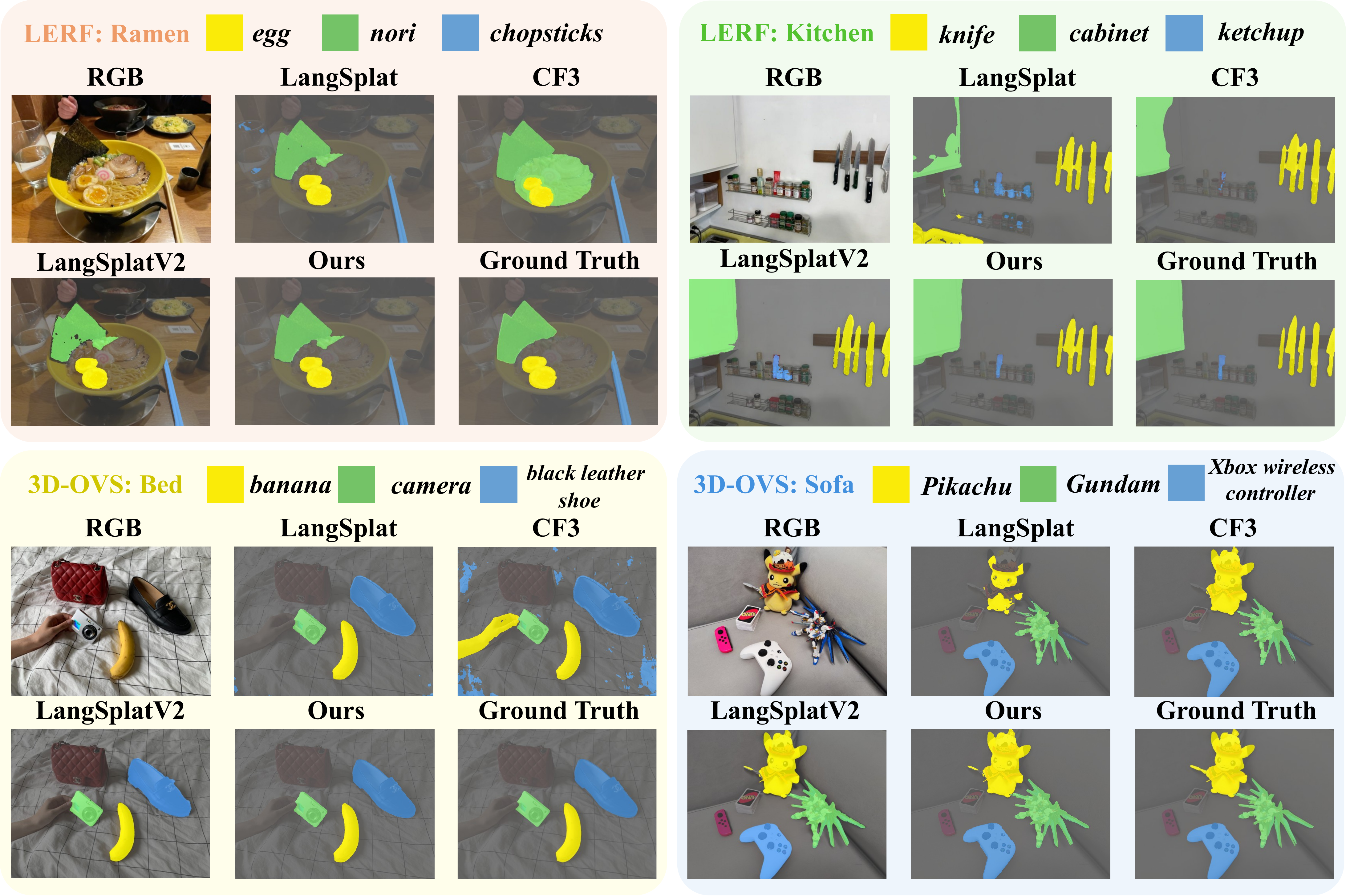}
\vspace{-14pt}
\caption{\textbf{Qualitative 2D segmentation on LERF and 3D-OVS.} CoSAG yields cleaner masks with
tighter boundaries and fewer spurious regions than LangSplat and CF3.}
\label{fig:2dseg}
\vspace{-15pt}
\end{figure}

Figure~\ref{fig:3dseclect} and Figure~\ref{fig:2dseg} present qualitative comparisons. In 2D
segmentation, CoSAG produces cleaner masks with tighter boundaries and fewer spurious regions than
LangSplat and SCOUP. In 3D selection, it isolates the queried object with coherent boundaries
where competing methods leak into the surrounding geometry. Since the anchors stay bound to the
original Gaussians, CoSAG further supports scene editing such as object removal, recoloring, and
duplication. Additional localization, Replica, editing, and large-scale Waymo~\cite{2020waymo} results appear in the \textbf{appendix}.

%% file: tables/table3.tex
\begin{table*}[t]
\centering
\caption{\textbf{Quantitative comparison for 3D object selection on LERF-OVS}. We report mAcc@0.25 (\%), mIoU (\%) and language-field storage (MB) per scene. \textcolor{red}{Red} and \textcolor{orange}{orange} highlight the best and second-best results in each column.}
\label{tab:3dsel}
\vspace{-5pt}
\resizebox{\textwidth}{!}{%
\begin{tabular}{l|ccc|ccc|ccc|ccc|ccc}
\toprule
\multirow{2}{*}{\textbf{Method}}
& \multicolumn{3}{c|}{\textbf{Teatime}} & \multicolumn{3}{c|}{\textbf{Ramen}}
& \multicolumn{3}{c|}{\textbf{Figurines}} & \multicolumn{3}{c|}{\textbf{Kitchen}}
& \multicolumn{3}{c}{\textbf{Overall}} \\
& Acc & mIoU & Stor. & Acc & mIoU & Stor. & Acc & mIoU & Stor. & Acc & mIoU & Stor.
& Acc & mIoU & Stor. \\
\midrule
LangSplat~\cite{langsplat} (CVPR'24)   & 8.47 & 7.91 & 27 & 5.63 & 3.49 & 9.8 & 14.29 & 9.64 & 10.5 & 9.09 & 9.61 & 26 & 9.37 & 7.66 & 18.3 \\
LEGaussians~\cite{legaussians} (CVPR'24)   & 33.90 & 21.93 & 72 & 28.17 & 14.15 & 27 & 25.00 & 16.40 & 28 & 27.27 & 17.21 & 70 & 28.56 & 17.42 & 49.3 \\
OpenGaussian~\cite{opengaussian} (NeurIPS'24) & 76.27 & 58.24 & 52 & 28.17 & 24.02 & 18 & 75.00 & \cellcolor{red!25}55.38 & 20 & 45.45 & 30.96 & 51 & 56.22 & 42.15 & 35.3 \\
Dr.Splat~\cite{drsplat} (CVPR'25)          & 76.27 & 57.20 & 275 & 35.21 & 24.70 & 95 & \cellcolor{red!25}80.36 & \cellcolor{orange!25}53.36 & 102 & \cellcolor{orange!25}63.64 & \cellcolor{orange!25}39.07 & 268 & 63.87 & 43.58 & 185.0 \\
LightSplat~\cite{lightsplat} (CVPR'26)     & \cellcolor{orange!25}79.66 & \cellcolor{orange!25}59.65 & \cellcolor{orange!25}4.6 & \cellcolor{red!25}57.75 & \cellcolor{red!25}45.07 & \cellcolor{orange!25}1.8 & \cellcolor{orange!25}76.79 & 50.63 & \cellcolor{orange!25}1.9 & 59.09 & 34.95 & \cellcolor{orange!25}4.5 & \cellcolor{orange!25}68.32 & \cellcolor{orange!25}47.58 & \cellcolor{orange!25}3.2 \\
\midrule
\textbf{CoSAG (Ours)}                      & \cellcolor{red!25}88.12 & \cellcolor{red!25}60.20 & \cellcolor{red!25}1.56 & \cellcolor{orange!25}52.14 & \cellcolor{orange!25}34.72 & \cellcolor{red!25}0.85 & 71.42 & 50.82 & \cellcolor{red!25}1.10 & \cellcolor{red!25}77.33 & \cellcolor{red!25}45.64 & \cellcolor{red!25}1.90 & \cellcolor{red!25}72.25 & \cellcolor{red!25}47.85 & \cellcolor{red!25}1.35 \\
\bottomrule
\end{tabular}}
\vspace{-8pt}
\end{table*}

%% file: tables/table4.tex
\begin{table}[t]
\centering
\caption{Dense semantic segmentation on Replica~\cite{straub2019replica} room\_0 under the
LSeg~\cite{lseg} protocol of DF-3DGS~\cite{df3dgs}. We report mIoU (\%), pixel accuracy (\%), and
semantic-field storage (MB); DF-3DGS storage includes its $4.05$\,MB shipped decoder.}
\label{tab:replica}
\setlength{\tabcolsep}{7pt}
\begin{tabular}{l|ccc}
\toprule
\textbf{Method} & mIoU & Pixel Acc & Stor. \\
\midrule
DF-3DGS (CVPR'25) & 79.07 & 90.85 & 8.31 \\
\textbf{CoSAG (Ours)}          & \textbf{80.19} & \textbf{91.15} & \textbf{0.34} \\
\bottomrule
\end{tabular}
\end{table}

%% file: tables/table5.tex

\begin{table}[t]
\centering
\caption{\textbf{Efficiency on LERF-OVS.} We compare construction time (hour), storage (MB),
query speed (FPS), and inference memory (GB) against representative baselines.}
\label{tab:eff}
\vspace{-3pt}
\setlength{\tabcolsep}{3.2pt}
\begin{tabular}{l|cccc}
\toprule
\textbf{Method} & Cons & Stor. & FPS & Mem. \\
\midrule
LangSplat~\cite{langsplat}      & 1.0 & 52.9 & 4.5 & 6.2 \\
LEGaussian~\cite{legaussians}   & 1.3 & 49.1 & 7.1 & 8.2 \\
CF3~\cite{cf3}                  & 0.8 & 4.2 & 26.4 & 6.5 \\
LangSplatV2~\cite{langsplatv2}  & 3.0 & 52.0 & 29.1 & 7.2 \\
\midrule
\textbf{CoSAG (Ours)}           & \textbf{0.7} & \textbf{1.4} & \textbf{30.2} & \textbf{6.0} \\
\bottomrule
\end{tabular}
\vspace{-5pt}
\end{table}

%% file: tables/table6.tex
\begin{table}[t]
\centering
\caption{Ablation of key components on LERF-OVS. mAcc (\%), mIoU (\%), storage (MB).}
\label{tab:ablation}
\vspace{-3pt}
\begin{tabular}{l|ccc}
\toprule
\textbf{Variant} & mAcc & mIoU & Stor. \\
\midrule
w/o multi-view averaging & 84.5 & 65.5 & 1.4 \\
Spatial-only binding     & 72.6 & 47.1 & 1.4 \\
Semantic-only binding    & 80.2 & 62.6 & \textbf{1.3} \\
\midrule
\textbf{Full (CoSAG)}    & \textbf{86.1} & \textbf{66.9} & 1.4 \\
\bottomrule
\end{tabular}
\vspace{-3pt}
\end{table}

%% file: tables/table7.tex
\begin{table}[t]
\centering
\caption{Binding coder ablation on the LERF dataset (all three levels): per-Gaussian bits
and total binding storage.}
\label{tab:coder}
\begin{tabular}{l|cc}
\toprule
\textbf{Coding stage} & bits/Gauss. & Stor. (MB) \\
\midrule
Raw int16                        & 48.0 & 8.65 \\
LZMA (no reorder)                & 31.4 & 5.78 \\
Morton reorder                   & 11.0 & 1.98 \\
parent-pointer sharing (Ours)    & \textbf{4.05} & \textbf{0.72} \\
\bottomrule
\end{tabular}
\vspace{-8pt}
\end{table}

%% file: section/appendix.tex
\appendix
\section{Appendix}
This appendix reports qualitative results and a large-scale study omitted from the main paper for
space. All results use the single fixed hyperparameter set of Sec.~4.1, with no per-scene tuning.
It is organized as follows:

\begin{itemize}
\item \textbf{Section A} shows additional 2D open-vocabulary localization on LERF.
\item \textbf{Section B} demonstrates 3D scene editing enabled by CoSAG.
\item \textbf{Section C} reports large-scale open-vocabulary understanding on Waymo.
\item \textbf{Section D} visualizes dense semantic segmentation on Replica.
\end{itemize}

\section{A~~Additional 2D Localization}
\begin{figure*}[t]
\centering
\includegraphics[width=1\textwidth]{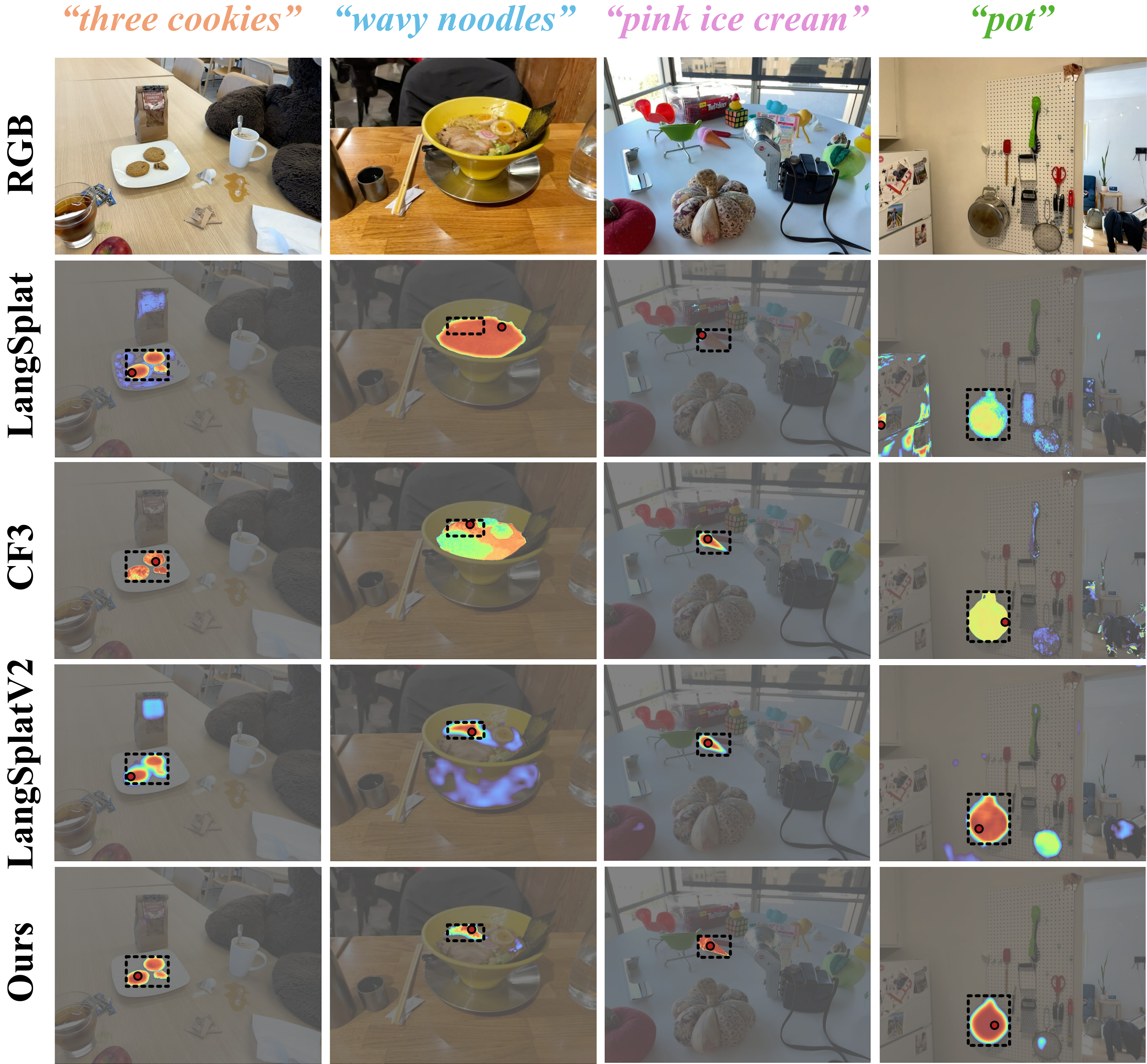}
\vspace{-5pt}
\caption{\textbf{Qualitative 2D localization on LERF.} CoSAG places a tighter, more focused
relevancy peak on the queried object than the baselines, with fewer spurious activations.}
\vspace{-5pt}
\label{fig:app_loc}
\end{figure*}

Figure~\ref{fig:app_loc} presents further open-vocabulary localization on LERF. Red points are the
predictions and dashed boxes the ground truth. CoSAG concentrates relevancy on the queried object
and localizes it more precisely, with fewer false peaks than LangSplat and SCOUP. This matches the
segmentation trend in the main paper and confirms that the denoised anchors give a sharp, focused
response even for small objects.

\section{B~~3D Scene Editing}
\begin{figure*}[t]
\centering
\includegraphics[width=1\textwidth]{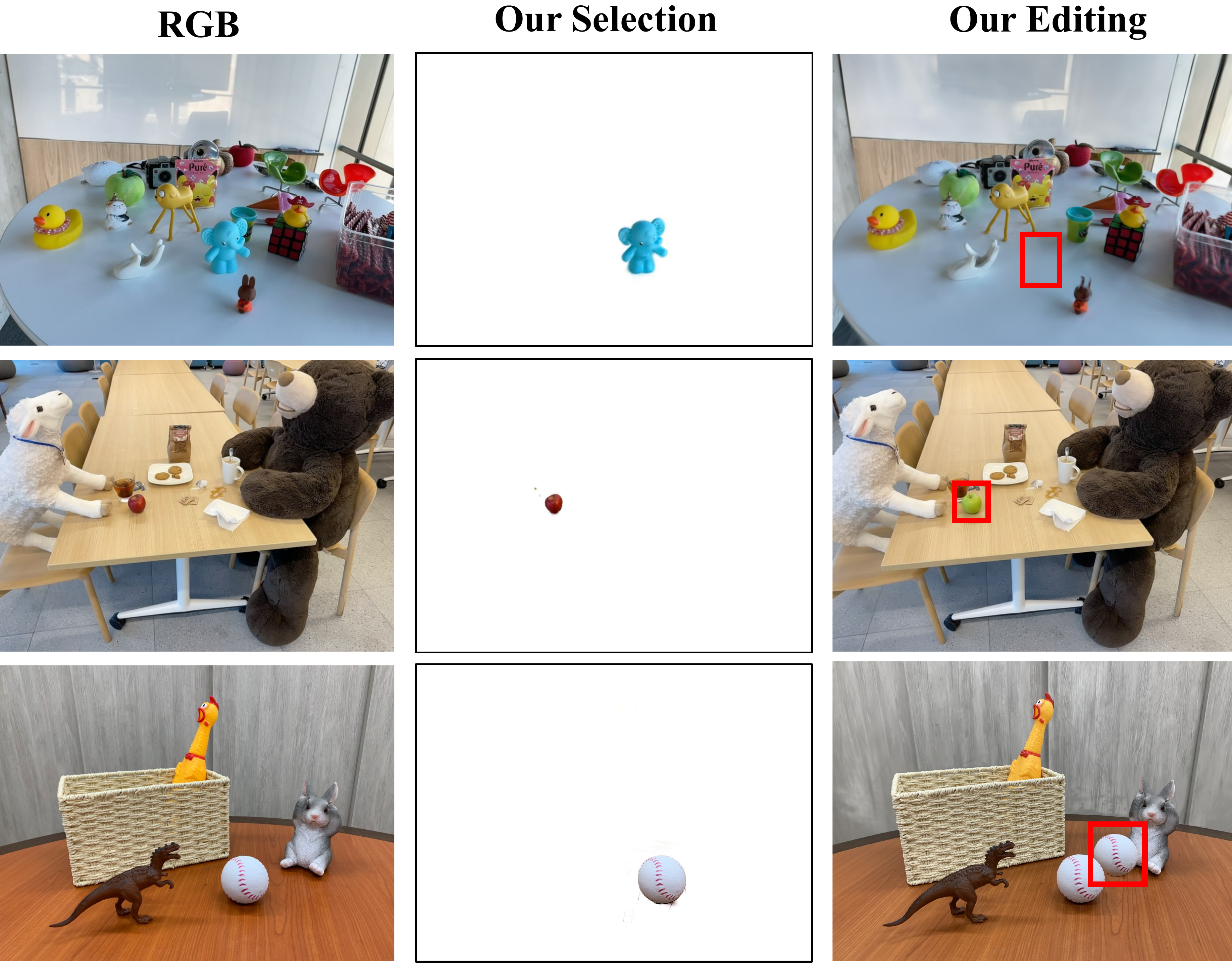}
\vspace{-5pt}
\caption{\textbf{3D scene editing enabled by CoSAG.} A text query selects an object in 3D and
supports removal, recoloring, and duplication, with the edited scene staying geometrically
consistent.}
\vspace{-5pt}
\label{fig:app_edit}
\end{figure*}

Because every anchor stays bound to a set of original Gaussians, a text query selects an object
directly in 3D, after which standard Gaussian operations edit it without any re-optimization.
Figure~\ref{fig:app_edit} shows object removal, recoloring, and duplication. The edited Gaussians
are exactly those bound to the matched anchors, so the untouched scene stays intact and the result
remains geometrically consistent. This is a direct benefit of keeping the semantics on the original
Gaussians rather than on a separate field.

\section{C~~Large-Scale Open-Vocabulary Understanding}
\begin{figure*}[t]
\centering
\includegraphics[width=1\textwidth]{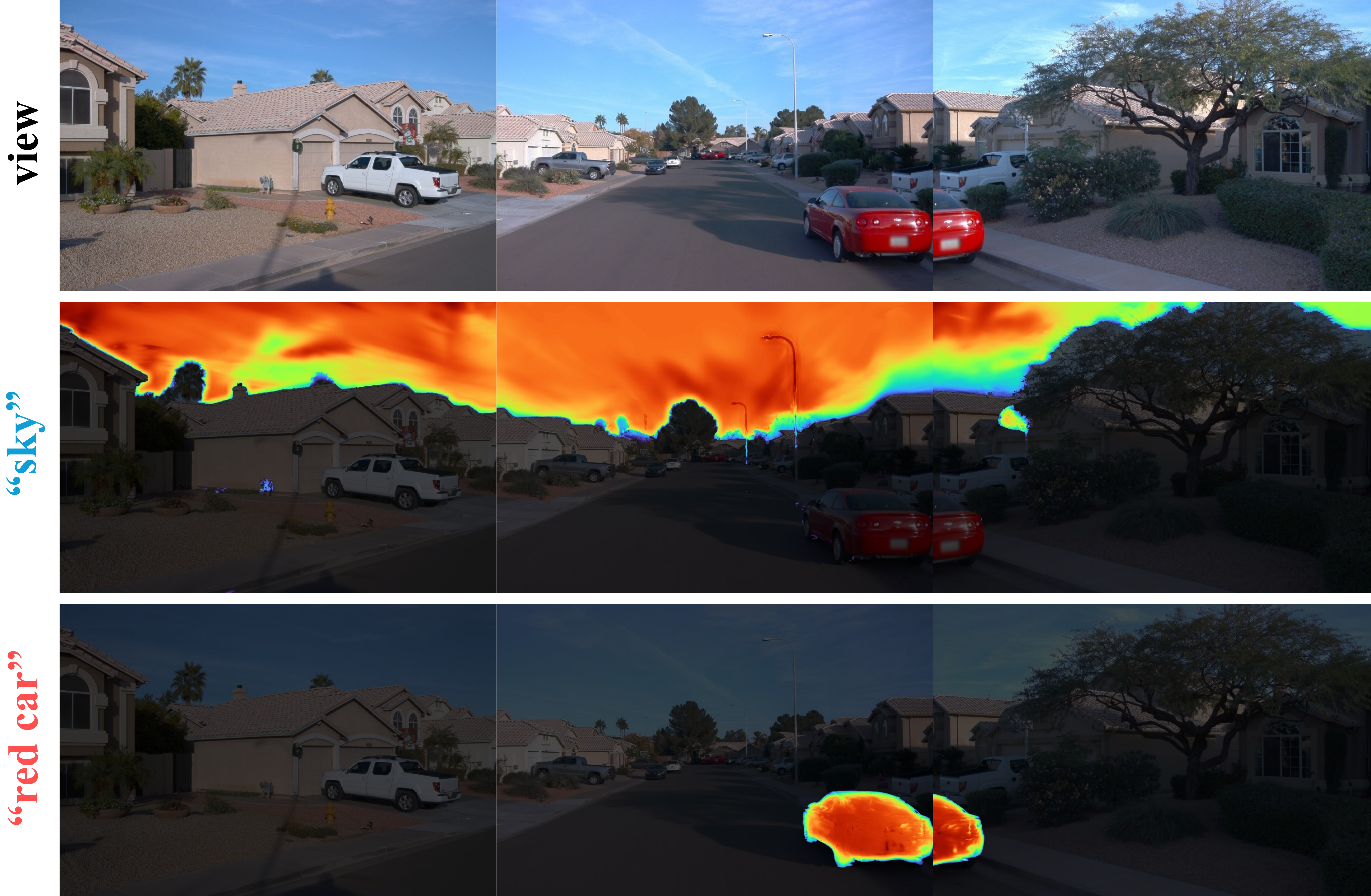}
\vspace{-5pt}
\caption{\textbf{Large-scale open-vocabulary query on Waymo.} CoSAG scales to an outdoor driving
scene with millions of Gaussians and keeps coherent relevancy for both large regions and small
objects.}
\vspace{-5pt}
\label{fig:app_waymo}
\end{figure*}

To test scalability beyond room-scale scenes, we reconstruct a large outdoor scene from the Waymo
Open Dataset. Following the setup used for indoor scenes, we take 180 images at
$960{\times}640$ from three front-facing cameras of segment-102751, giving a reconstruction with up
to $2{,}500$K Gaussians. CoSAG constructs and compresses the semantic field with the same fixed
hyperparameters, without any scene-specific tuning. As Figure~\ref{fig:app_waymo} shows, it yields
coherent relevancy for both large regions such as the road and buildings and small objects such as
vehicles and signs, while storing the field in only 2.6 MB. The training-free
construction and rate-distortion storage thus scale to millions of Gaussians without change.

\section{D~~Dense Semantic Segmentation on Replica}
\begin{figure*}[t]
\centering
\includegraphics[width=1\textwidth]{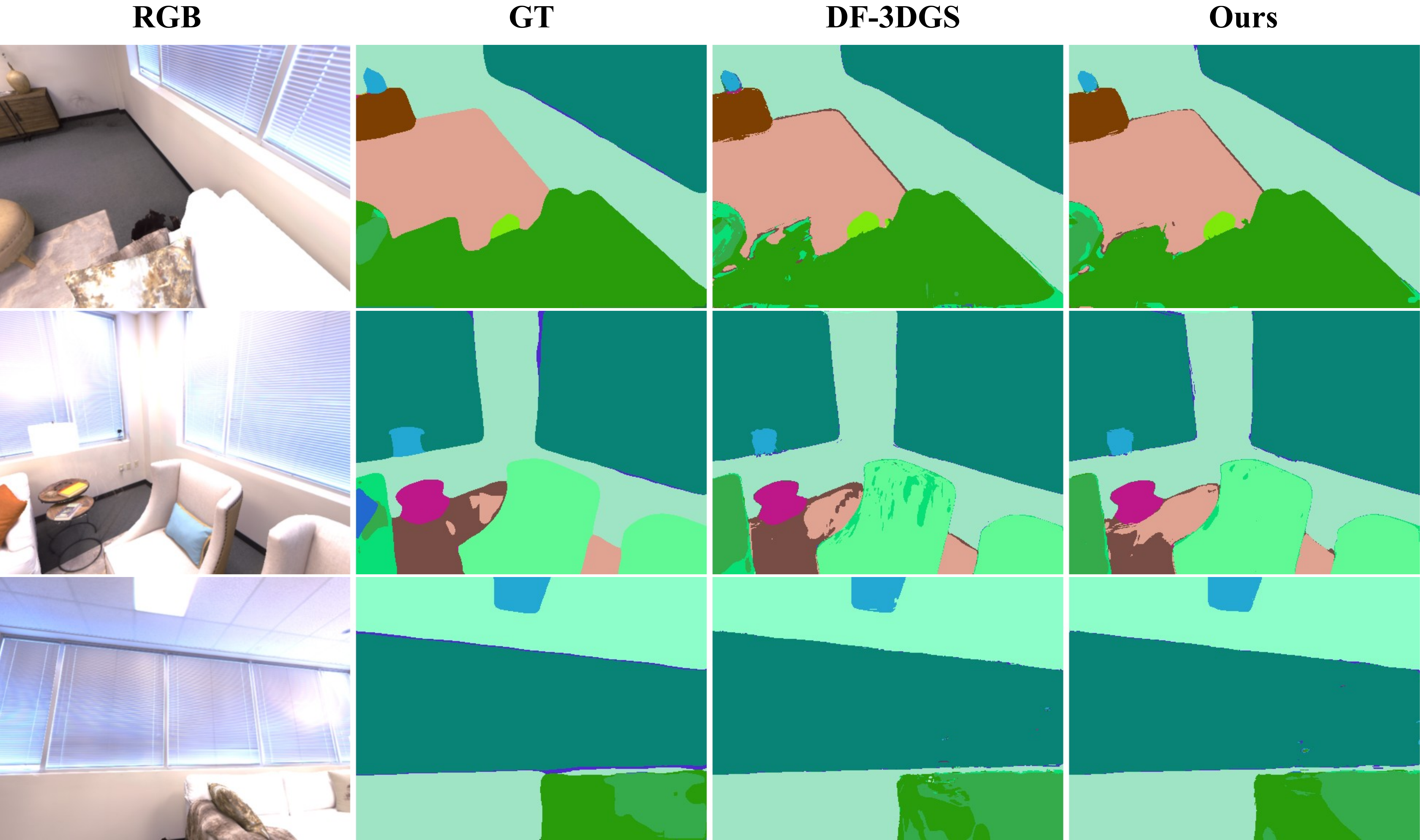}
\vspace{-5pt}
\caption{\textbf{Dense semantic segmentation on Replica room\_0.} Under the LSeg protocol, CoSAG
reconstructs the dense semantic field faithfully while storing it in only $0.34$\,MB.}
\vspace{-5pt}
\label{fig:app_replica}
\end{figure*}

Figure~\ref{fig:app_replica} visualizes the dense LSeg segmentation on Replica room\_0 discussed in
Sec.~4.4. CoSAG recovers the per-pixel semantics faithfully across the scene while storing the field
in $0.34$\,MB, far below DF-3DGS. The reconstruction stays sharp at object boundaries, confirming
that the closed-form lift and shared-basis coding preserve dense semantics as well as SAM-region
CLIP features.

